\documentclass{article}
\usepackage[final, nonatbib]{neurips_2018}

\usepackage{amsmath, amssymb, amsthm}
\DeclareMathOperator*{\argmax}{arg\,max}
\DeclareMathOperator*{\argmin}{arg\,min}

\usepackage[utf8]{inputenc} 
\usepackage[T1]{fontenc}    
\usepackage{hyperref}       
\usepackage{url}            
\usepackage{booktabs}       
\usepackage{amsfonts}       
\usepackage{nicefrac}       
\usepackage{microtype}      

\usepackage{verbatim}
\usepackage{adjustbox}
\usepackage{stmaryrd}
\usepackage{epigraph}
\usepackage{graphicx}
\usepackage{wrapfig}
\usepackage{example_sentences}
\usepackage{tikz}
\tikzset{%
	neuron/.style={
		circle, draw, minimum size=0.25cm
	},
}

\usepackage{biblatex}
\title{Paying Attention to Function Words}
\author{Shane Steinert-Threlkeld\thanks{Thanks to Jeff Barrett, Emmanuel Chemla, Meica Magnani, Iris van de Pol, and Jakub Szymanik for helpful comments and discussion.  An extended version of this paper was presented at the Workshop on Evolutionary Explanations of Compositional Communication at the Biennial Conference of the Philosophy of Science Association: see~\url{http://philsci-archive.pitt.edu/15274/}.  This work  was  supported  by  funding  from  the  European  Research  Council  under  the  European  Union’s Seventh Framework Programme (FP/2007-2013)/ERC Grant Agreement n.  STG 716230 CoSaQ.}
	\\ Institute for Logic, Language and Computation, Universiteit van Amsterdam \\ \texttt{S.N.M.Steinert-Threlkeld@uva.nl}}

\begin{document}

\maketitle

\epigraph{
	\emph{Twas} brillig, \emph{and the} slithy toves

	\emph{Did} gyre \emph{and} gimble \emph{in the} wabe;

	\emph{All} mimsy \emph{were the} borogoves,

	\emph{And the} mome raths outgrabe.	
}{Excerpt from `Jabberwocky' in \textcite{Carroll1871}.}

The poem excerpted in the epigraph has often been called a `nonsense poem'.  But it is not entirely so.  While the \emph{content} words (unemphasized: nouns, verbs, adjectives) are nonsense, the \emph{function} words (emphasized: determiners, tense, auxiliaries, conjunctions, etc.) are not.  The structure that they provide greatly aids our interpretation.

The distinction between these two types of expression occupies a central place in modern linguistics \cite{Carnie2006, Muysken2008, Rizzi2016}.  Rightfully so: every natural language exhibits a distinction between content and function words.  The former provide the content of sentences and fall into what are called `open classes' (it is easy to introduce a new noun, for instance) while the latter provide the `grammatical glue' of complex expressions and fall into `closed classes' (it is difficult to introduce a new determiner, for instance).

Yet surprisingly little has been said about the emergence of this universal architectural feature of natural languages.  Why have human languages evolved to exhibit this division of labor between content and function words?  How could such a distinction have emerged in the first place?

This paper takes steps towards answering these questions by showing how the distinction can emerge through reinforcement learning in agents playing a signaling game across contexts which contain multiple objects that possess multiple perceptually salient gradable properties.  In the next section, I will introduce the new signaling game.  Section~\ref{exp} presents experimental results.  After discussing related work in Section~\ref{related}, I conclude with future directions in Section~\ref{conclusion}.

\section{A Signaling Game with Varying Contexts}
\label{game}

I will introduce a type of signaling game \cite{Lewis1969, Skyrms2010} -- called the Extremity Game -- with a few helper definitions.  Following the literature on gradable adjectives \cite{Kennedy2005, Kennedy2007}, I will assume that objects have some number of gradable properties, where each property has a corresponding \emph{scale}.  A scale in turn is a set of \emph{degrees}, totally ordered with respect to a dimension.  For example, the size of a circle corresponds to its radius, with degrees being positive real numbers (i.e.\ $\mathbb{R}^+$).  For the degree of an object $o$ on a scale $s$, I will write $s(o)$.  Given a set $S$ of scales, I will define a context as follows.
\begin{examples}
\item A \emph{context} $c$ over scales $S$ is a set of objects such that: for each $o \in c$, there is a scale $s \in S$ such that either $o$ has the least degree on $s$ ($o = \argmin_{o' \in c} s(o')$) or the highest degree on $s$ ($o = \argmax_{o' \in c} s(o')$). \label{context}
\end{examples}
At its most general form, the game takes place between a sender and a receiver in the following way.
\begin{examples}
\item Extremity Game, in general: \label{extremity-game}
	\begin{examples}
	\item Nature chooses a context $c$ and a target object $o \in c$.
	\item The sender sees $c$ and $o$ and sends a message $m$ from some set of messages $M$.
	\item The receiver sees $c$ and $m$ and chooses an object $o'$ from $c$.
	\item The play is successful (and the two agents equally rewarded) if and only if $o' = o$.
	\end{examples}
\end{examples}

To fully specify a game, one must say what the messages $M$ available are and how the agents make their choices.  I will specify the former now and the latter in the next section.  The set of available messages will be inspired by the semantics for gradable adjectives.  There, it is assumed that adjectives map objects (of type $e$) on to their degree on the corresponding scale (of type $d$).  Morphemes like \emph{-est} and \emph{least} then map a contextually specified set of objects to those with the highest and lowest degrees, respectively.
\begin{examples}
\item Toy semantics for a gradable adjective and superlative morphemes. \label{gradablesemantics}
	\begin{examples}
	\item $\llbracket \text{size} \rrbracket = \lambda x . s_\text{size}(x)$
	\item $\llbracket \text{-est} \rrbracket^c = \lambda P_{\langle e, d \rangle} . \lambda x_e . x \in c \text{ and } \forall x' \in c , P(x) \succeq P(x')$
	\item $\llbracket \text{least} \rrbracket^c = \lambda P_{\langle e, d \rangle} . \lambda x_e . x \in c \text{ and } \forall x' \in c , P(x) \preceq P(x')$
	\end{examples}
\end{examples}
In contexts as defined in \exref{context}, having one expression for each scale and the morphemes \emph{-est} and \emph{least} will suffice to uniquely pick out each object in the context.  I will assume, then, that the set of messages $M = M_S \times M_P$ where $M_S$ is a set of size $|S|$ (i.e.\ there are as many messages in $M_S$ as there are gradable properties for each object) and $M_P$ is a set of size two ($P$ for `polarity').  

\section{Experiment}
\label{exp}

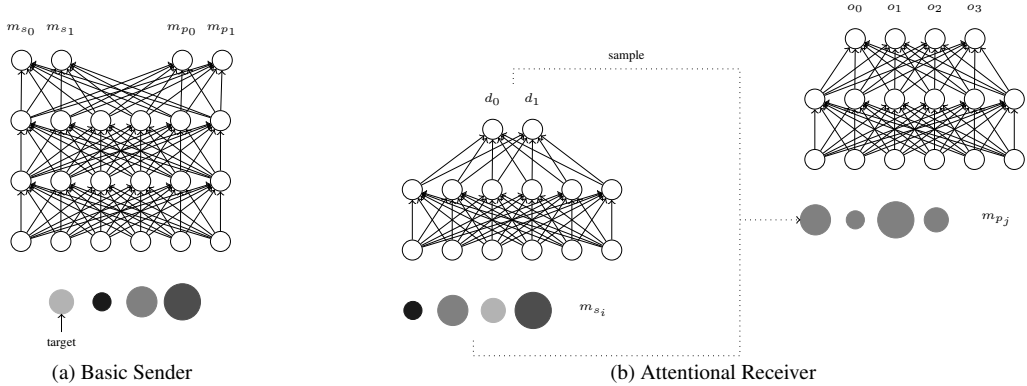
\begin{figure}[ht]
	\centering
	\begin{adjustbox}{max width=\textwidth}
	\begin{tabular}{cc}
		\begin{tikzpicture}
		\draw[fill, opacity=0.3] (-1, 0) circle [radius=0.2cm];
		\draw[fill, opacity=0.9] (-0.33, 0) circle [radius=0.15cm];
		\draw[fill, opacity=0.5] (0.33, 0) circle [radius=0.25cm];
		\draw[fill, opacity=0.7] (1, 0) circle [radius=0.3cm];
		\draw[->] (-1, -0.5) -- (-1, -0.2);
		\node at (-1, -0.7) {\tiny target};
		\foreach \m [count=\y] in {0, 1, 2, 3, 4, 5}
			\node[neuron] (input-\m) at (-2.33+0.66*\y, 1) {};
		\foreach \m [count=\y] in {0, 1, 2, 3, 4, 5}
			\node[neuron] (h1-\m) at (-2.33+0.66*\y, 2) {};
		\foreach \m [count=\y] in {0, 1, 2, 3, 4, 5}
			\node[neuron] (h2-\m) at (-2.33+0.66*\y, 3) {};
		\node[neuron] (m0-0) at (-1.66, 4) {};
		\node[neuron] (m0-1) at (-1, 4) {};
		\node at (-1.66, 4.5) {\tiny $m_{s_0}$};
		\node at (-1, 4.5) {\tiny $m_{s_1}$};
		\node[neuron] (m1-0) at (1, 4) {};
		\node[neuron] (m1-1) at (1.66, 4) {};
		\node at (1, 4.5) {\tiny $m_{p_0}$};
		\node at (1.66, 4.5) {\tiny $m_{p_1}$};
		\foreach \i in {0, 1, 2, 3, 4, 5}
			\foreach \j in {0, 1, 2, 3, 4, 5}
				\draw[->] (input-\i) -- (h1-\j);
		\foreach \i in {0, 1, 2, 3, 4, 5}
			\foreach \j in {0, 1, 2, 3, 4, 5}
				\draw[->] (h1-\i) -- (h2-\j);
		\foreach \i in {0, 1, 2, 3, 4, 5}
			\foreach \j in {0, 1}
				\draw[->] (h2-\i) -- (m0-\j);
		\foreach \i in {0, 1, 2, 3, 4, 5}
			\foreach \j in {0, 1}
				\draw[->] (h2-\i) -- (m1-\j);
		\end{tikzpicture}
		\hspace{1cm}
		&
		\hspace{1cm}
		\begin{tikzpicture}
		\draw[fill, opacity=0.3] (-0.33, 0) circle [radius=0.2cm];
		\draw[fill, opacity=0.9] (-1.66, 0) circle [radius=0.15cm];
		\draw[fill, opacity=0.5] (-1, 0) circle [radius=0.25cm];
		\draw[fill, opacity=0.7] (0.33, 0) circle [radius=0.3cm];
		\node at (1.33, 0) {\tiny $m_{s_i}$};
		\foreach \m [count=\y] in {0, 1, 2, 3, 4, 5}
			\node[neuron] (input-\m) at (-2.33+0.66*\y, 1) {};
		\foreach \m [count=\y] in {0, 1, 2, 3, 4, 5}
			\node[neuron] (h1-\m) at (-2.33+0.66*\y, 2) {};
		\foreach \m [count=\y] in {0, 1}
			\node[neuron] (dim-\m) at (-1+0.66*\y, 3) {};
		\foreach \m [count=\y] in {0, 1}
			\node at (-1+0.66*\y, 3.5) {\tiny $d_\m$};
		\foreach \i in {0, 1, 2, 3, 4, 5}
			\foreach \j in {0, 1, 2, 3, 4, 5}
				\draw[->] (input-\i) -- (h1-\j);
		\foreach \i in {0, 1, 2, 3, 4, 5}
			\foreach \j in {0, 1}
				\draw[->] (h1-\i) -- (dim-\j);

		\draw[fill, opacity=0.5] (7, 1.5) circle [radius=0.2cm];
		\draw[fill, opacity=0.5] (5.66, 1.5) circle [radius=0.15cm];
		\draw[fill, opacity=0.5] (5, 1.5) circle [radius=0.25cm];
		\draw[fill, opacity=0.5] (6.33, 1.5) circle [radius=0.3cm];
		\node at (8, 1.5) {\tiny $m_{p_j}$};
		\foreach \m [count=\y] in {0, 1, 2, 3, 4, 5}
			\node[neuron] (input-\m) at (4.33+0.66*\y, 2.5) {};
		\foreach \m [count=\y] in {0, 1, 2, 3, 4, 5}
			\node[neuron] (h1-\m) at (4.33+0.66*\y, 3.5) {};
		\foreach \m [count=\y] in {0, 1, 2, 3}
			\node[neuron] (obj-\m) at (5+0.66*\y, 4.5) {};
		\foreach \m [count=\y] in {0, 1, 2, 3}
			\node at (5+0.66*\y, 5) {\tiny $o_\m$};
		\foreach \i in {0, 1, 2, 3, 4, 5}
			\foreach \j in {0, 1, 2, 3, 4, 5}
				\draw[->] (input-\i) -- (h1-\j);
		\foreach \i in {0, 1, 2, 3, 4, 5}
			\foreach \j in {0, 1, 2, 3}
				\draw[->] (h1-\i) -- (obj-\j);

		\draw[dotted] (0, 3.75) -- (0, 4) -- node[above] {\tiny sample} (3.75, 4) -- (3.75, 1.5);
		\draw[dotted] (-0.66, -0.5) -- (-0.66, -0.75) -- (3.75, -0.75) -- (3.75, 1.5);
		\draw[->, dotted] (3.75, 1.5) -- (4.75, 1.5);
		\end{tikzpicture}
	
		\\
		(a) Basic Sender 
		\hspace{1cm}
		& 
		\hspace{1cm}
		(b) Attentional Receiver
	\end{tabular}
	\end{adjustbox}
	\caption{Schematic depictions of network architectures.}  
	\label{fig:networks-basic}
\end{figure}

A trial of our experiment will consist of some number of iterations of playing an Extremity Game as in \exref{extremity-game}.  The sender and receiver are each neural networks, schematically depicted in Figure~\ref{fig:networks-basic}.  They are trained using the REINFORCE algorithm (\cite{Williams1992, Sutton2018}).  
There are two types of receiver: Basic and Attentional.  The Basic one is a multilayer perceptron, taking the context and the signals, outputting a distribution over target objects, from which a sample is taken to determine the reward.

The Attentional Receiver uses an \emph{attention mechanism} (\cite{Mnih2014, Xu2015}) to focus on a perceptually salient dimension.  They implement a \emph{hard} attention mechanism in the following sense.  First, they receive as input the context $c$ and the message $m_{s_i}$ from $M_S$ chosen by the sender.  On this basis, the receiver \emph{chooses a dimension to attend to}: the input is filtered so that the agent only sees the objects according to one dimension (e.g.\ size or lightness).  Then, the agent uses this attended-to dimension and the message from $M_P$ chosen by the sender to choose a target object.  This attention mechanism reflects the perceptual salience of the gradable properties of the objects: it is very natural, for instance, in the contexts depicted in Figure~\ref{fig:networks-basic}, to attend only to the size or the shade of the circles.

I varied the number of dimensions (i.e.\ gradable properties) between 1 and 3, and ran 10 trials for each (for five-, twenty-, and fifty-thousand mini-batches respectively, where each mini-batch was size 64). I recorded the rolling accuracy over 10 training steps, as well as the accuracy and detailed properties about contexts and signals used on 5000 new games at the end of training.  Complete details of the architecture and training set-up, as well as a link to the code, are included in an Appendix. 

\paragraph{Results: Basic Receiver}

Mean communicative success per number of dimensions on 5000 novel games is provided in Table~\ref{tab:acc_base}.  In one and two dimensions, the agents reliably learned to communicate effectively.  In three dimensions, they usually get stuck in sub-optimal protocols.
\begin{wraptable}{R}{0pt}
	\centering
	\begin{tabular}{ccc}
		\toprule
		dims & mean & std
		\\
		\midrule
		1 & 0.975 & 0.006 \\
		2 & 0.985 & 0.003 \\
		3 & 0.731 & 0.062 \\
		\bottomrule
	\end{tabular}
	\caption{Success on test, Basic Receiver.}
	\vspace{-0.8em}
	\label{tab:acc_base}
\end{wraptable}

Inspection of the learned communication protocols also show that they do not learn to treat either of the signals as a function word.
The learned systems are always `maximally' separating in the following sense: for any two contexts $c, c'$ and targets $o, o'$, if $o = \argmin_c s_d(o)$ and $o' = \argmax_{c'} s_d(o)$ for the same dimension $d$, then the sender's message for $o$ in $c$ differs from its message for $o'$ in $c'$ in both syntactic positions.  This holds true for both the 2- and 3-dimensional cases.  In such a system, the agents are not grouping context/target pairs according to the dimension along which the target can be singled out as maximal or minimal.

This could be for roughly the following reason: in expectation, target objects that differ only in whether they are the minimum/maximum in context on the same dimension will actually be farther from each other in Euclidean space than from other objects.  So the sender could be using maximally different signals for the two types of target objects to help the receiver distinguish them.

\paragraph{Results: Attentional Receiver}

Mean communicative success per number of dimensions on 5000 novel games is provided in Table~\ref{tab:acc_att}.  In one and two dimensions, the agents reliably learned to communicate effectively.  In three dimensions, we find a lower mean and higher variance.  Visual inspection shows that many trials wind up near 88\% communicative success, while others get stuck in very sub-optimal communication protocols.
\begin{wraptable}{R}{0pt}
	\centering
	\begin{tabular}{ccc}
		\toprule
		dims & mean & std
		\\
		\midrule
		1 & 0.959 & 0.005 \\
		2 & 0.964 & 0.005 \\
		3 & 0.697 & 0.144 \\
		\bottomrule
	\end{tabular}
	\caption{Success on test, Attentional Receiver.}
	\vspace{-0.8em}
	\label{tab:acc_att}
\end{wraptable}

\begin{figure}[ht]
	\centering
	\begin{tabular}{cc}
	\includegraphics[width=0.5\textwidth]{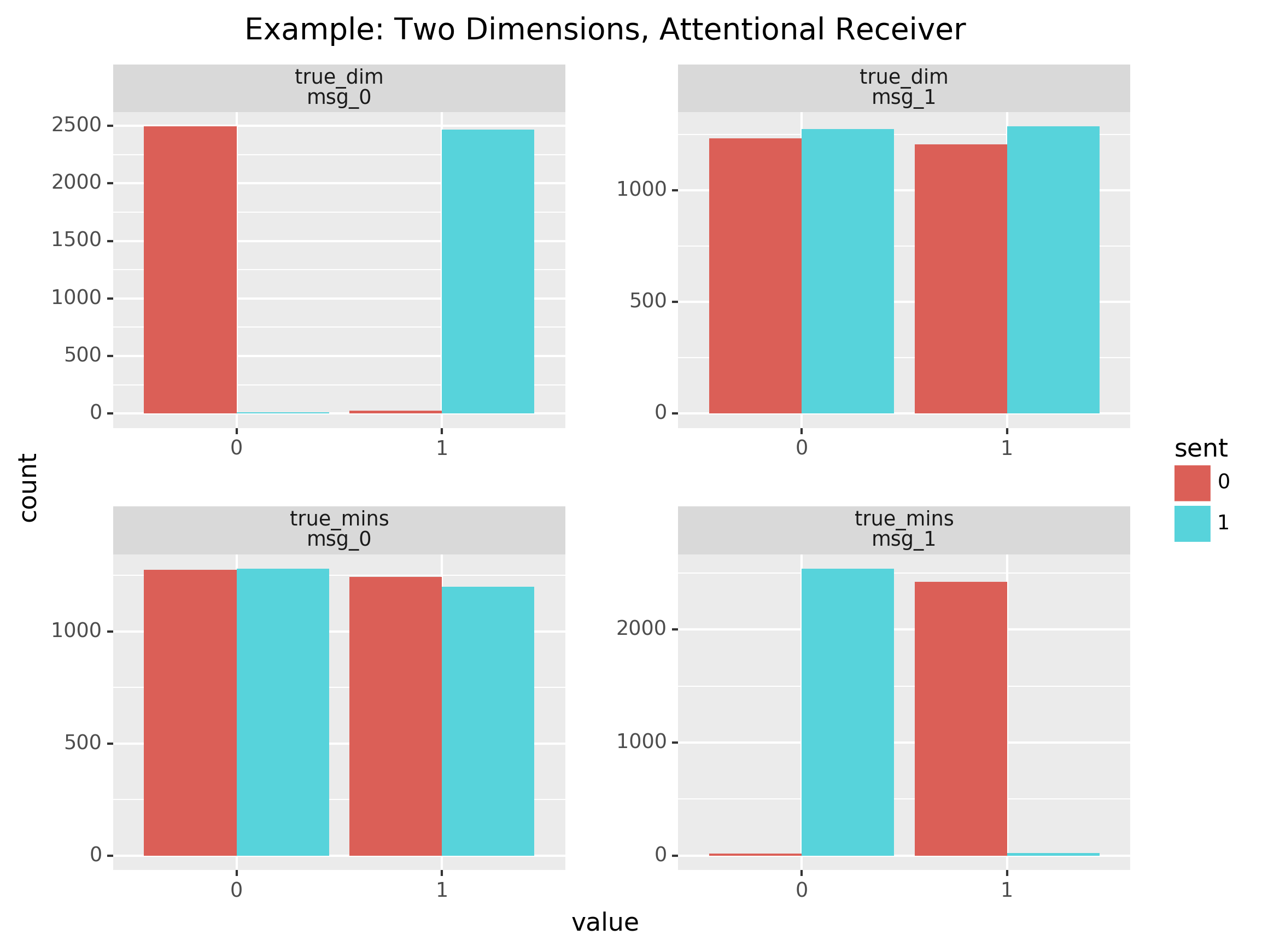}
	& 
	\includegraphics[width=0.5\textwidth]{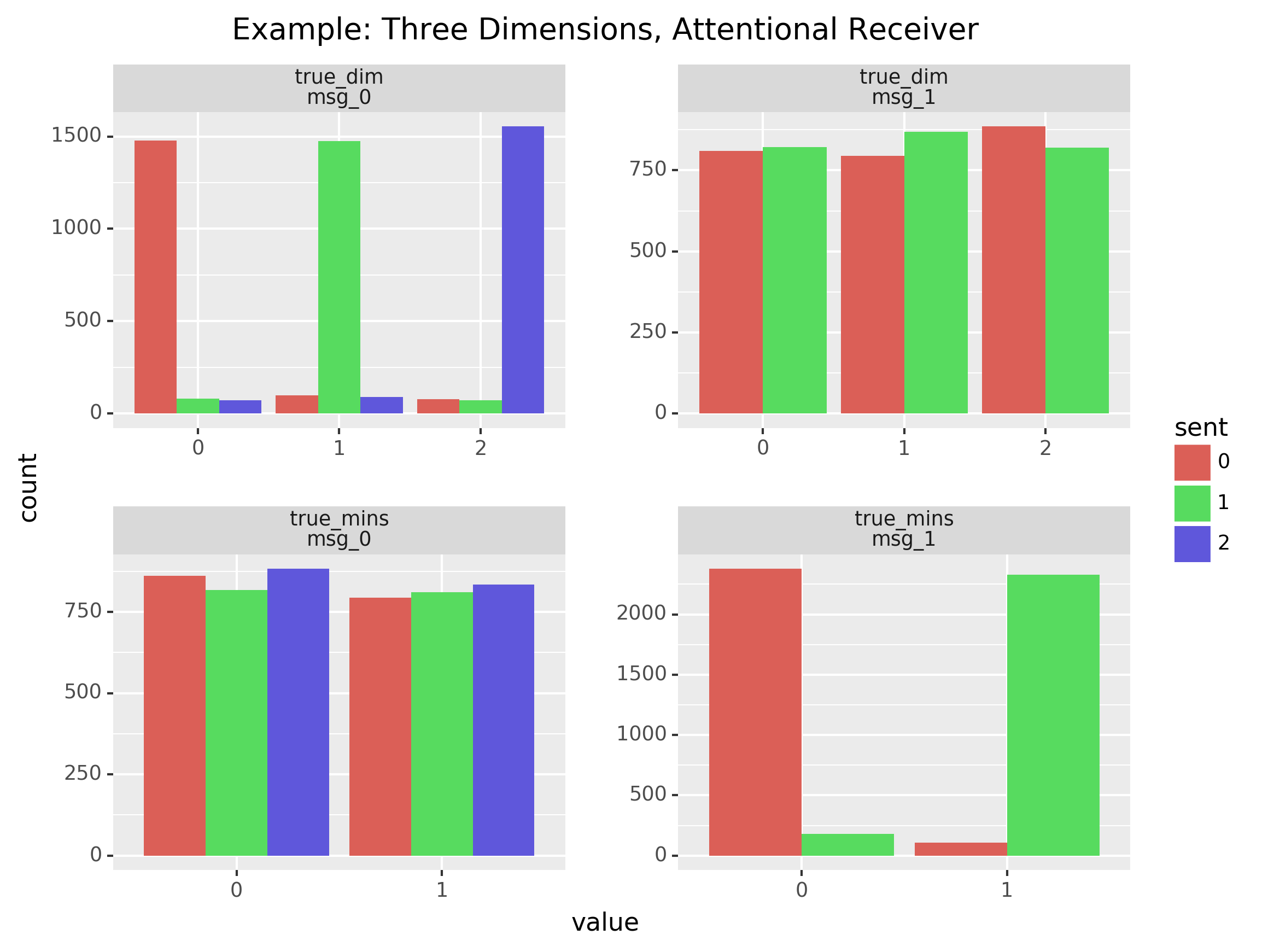}
	\end{tabular}
	\caption{Example communication systems with two and three dimensions.}
	\label{fig:attlang}
\end{figure}

Analyzing the resulting communication protocols yields promising results.  Figure~\ref{fig:attlang} shows an example learned communication system for a two-dimension (left) and three-dimension (right) trial.  These are bar plots, showing the frequency with which the sender made various choices on the test games. The left column corresponds to $M_S$, and the right to $M_P$.  The top row corresponds to the true dimension of the target object in context, and the bottom row to the true polarity of the target object.  The top-left corner in each case shows that the different signals in $M_S$ are being used to reliably communicate the dimension.  The bottom-right corner in each figure shows that the signals in $M_P$ are reliably being used to communicate the polarity of the object.

When the agents are communicating in this way, the signals that communicate direction can be interpreted as function words.  The signals in $M_S$ reliably communicate a bit of `content': a dimension.  The signals in $M_P$ reliably signal whether the target has the greatest/lowest degree \emph{along that dimension} of all the objects in the context.  This is non-trivial modification of one linguistic item by another.  The resulting communication protocols behave exactly like the toy semantics in \exref{gradablesemantics}. 

\section{Related Work}
\label{related}

Three prominent studies of the evolution of compositionality from different disciplines have yielded much insight, but do not explain the emergence of function words.  \textcite{Nowak1999} show that an evolutionary dynamic will eventually lead to the adoption of a `grammatical' language: messages here are pairs of signals, with one signal corresponding to an object (noun) and the other to an action (verb).  Such a language has no function words.  \textcite{Barrett2007a, Barrett2009} studies reinforcement learning in simple signaling games with multiple senders.  Here, the senders learn to jointly partition the state space, and the receiver learns to interpret by set intersection.  All of the signals provide content, which the receiver then interprets via a simple conjunctive operation.  \textcite{Mordatch2018} apply multi-agent reinforcement learning to agents who desire to get other agents to perform certain actions on certain landmarks in their shared environment.  In the resulting systems, there are separate signals for each agent, type of action, and landmark; the complex signals are interpreted `conjunctively', without sensitivity to word order.

In the extended version of this paper, linked in the acknowledgments footnote on the first page, I prove a limitative result that explains why these and similar models fail to account for the emergence of function words.  Roughly: if optimal communication consists in recovering from a sequence of symbols a target object from among a fixed set of states, the receiver interprets complex signals as generalized conjunction.  In such systems (which are there called trivially compositional), there are no function words.\footnote{For examples of systems that avoid the assumptions of the result, see \cite{Barrett2018, SteinertThrelkeld2016}.}

\textcite{Lazaridou2017} develop an approach to learning in reference games which directly inspired the present one.  Their contexts consist of two natural images, one of which is the target.  The sender chooses one signal from a fixed-sized vocabulary to send to the receiver.  While they are interested in whether natural concepts emerge in such a setting, I am focused on less natural input but more complex communication protocols in order to explore the emergence of functional vocabulary.

\section{Conclusion}
\label{conclusion}

Every natural language divides the lexicon into content and function words.  The latter provide the `grammatical glue' that enables robust forms of compositional communication to arise.  Most existing approaches to the evolution of compositionality do not explain the emergence of function words.  In this paper, I introduced a signaling game with variable contexts consisting of multiple objects with multiple gradable properties.  Simple reinforcement learning by neural networks -- in particular with the ability to pay attention to certain perceptually salient aspects of the input -- in this game can generate expressions that are appropriately characterized as function and as content words.

Much work remains to be done.  One would like architectures that make fewer assumptions about what aspects of the input the receiver pays attention to.  A first step in this direction will be to use a soft, as opposed to hard, attention mechanism.  A more thorough hyper-parameter search may also generate more reliable learning results in the higher-dimensional settings.  One can also generalize the input so that the networks additionally have to discover \emph{which dimensions} are relevant for being able to successfully refer to objects across contexts, instead of having it built into the current definition of context that every dimension is in principle useful in every context.
More generally, one would like communication systems like those exhibited here to emerge in the very general setting of communicating by a sequence of symbols (e.g. with recurrent neural network senders and receivers), with costs for things like vocabulary size and length of messages.
All of these exciting avenues remain to be pursued in future work.

\printbibliography

@book{Muysken2008,
address = {Cambridge},
author = {Muysken, Pieter},
publisher = {Cambridge University Press},
title = {{Functional Categories}},
year = {2008}
}

@article{Rizzi2016,
author = {Rizzi, Luigi and Cinque, Guglielmo},
doi = {10.1146/annurev-linguistics-011415-040827},
journal = {Annual Review of Linguistics},
number = {1},
pages = {139--163},
title = {{Functional Categories and Syntactic Theory}},
volume = {2},
year = {2016}
}

@book{Carnie2006,
address = {Oxford},
author = {Carnie, Andrew},
edition = {Second},
publisher = {Blackwell Publishing},
title = {{Syntax: A Generative Introduction}},
year = {2006}
}

@book{Sutton2018,
author = {Sutton, Richard S and Barto, Andrew G},
edition = {Second Edi},
publisher = {The MIT Press},
title = {{Reinforcement learning: an introduction.}},
year = {2018}
}

@book{Skyrms2010,
author = {Skyrms, Brian},
publisher = {Oxford University Press},
shorttitle = {Signals},
title = {{Signals: Evolution, Learning, and Information}},
year = {2010}
}

@article{Mnih2014,
archivePrefix = {arXiv},
arxivId = {1406.6247},
author = {Mnih, Volodymyr and Heess, Nicolas and Graves, Alex and Kavukcuoglu, Koray},
eprint = {1406.6247},
pages = {1--12},
title = {{Recurrent Models of Visual Attention}},
url = {http://arxiv.org/abs/1406.6247},
year = {2014}
}

@article{Ioffe2015,
archivePrefix = {arXiv},
arxivId = {1502.03167},
author = {Ioffe, Sergey and Szegedy, Christian},
eprint = {1502.03167},
title = {{Batch Normalization: Accelerating Deep Network Training by Reducing Internal Covariate Shift}},
url = {http://arxiv.org/abs/1502.03167},
year = {2015}
}

@inproceedings{Lazaridou2017,
archivePrefix = {arXiv},
arxivId = {1612.07182},
author = {Lazaridou, Angeliki and Peysakhovich, Alexander and Baroni, Marco},
booktitle = {International Conference of Learning Representations (ICLR2017)},
eprint = {1612.07182},
title = {{Multi-Agent Cooperation and the Emergence of (Natural) Language}},
url = {http://arxiv.org/abs/1612.07182},
year = {2017}
}

@inproceedings{Mordatch2018,
author = {Mordatch, Igor and Abbeel, Pieter},
booktitle = {The Thirty-Second AAAI Conference on Artificial Intelligence (AAAI 2018)},
title = {{Emergence of Grounded Compositional Language in Multi-Agent Populations}},
url = {http://arxiv.org/abs/1703.04908},
year = {2018}
}

@inproceedings{Barrett2018,
author = {Barrett, Jeffrey A and Skyrms, Brian and Cochran, Calvin},
booktitle = {26th Philosophy of Science Association Biennnial Meeting},
title = {{Hierarchical Models for the Evolution of Compositional Language}},
year = {2018}
}

@book{Lewis1969,
author = {Lewis, David},
publisher = {Blackwell},
title = {{Convention}},
year = {1969}
}

@article{Nowak1999,
author = {Nowak, Martin A and Krakauer, David C},
journal = {Proceedings of the National Academy of Sciences},
pages = {8028--8033},
title = {{The evolution of language}},
volume = {96},
year = {1999}
}

@article{SteinertThrelkeld2016,
author = {Steinert-Threlkeld, Shane},
doi = {10.1007/s10849-016-9236-9},
journal = {Journal of Logic, Language and Information},
number = {3},
pages = {379--397},
title = {{Compositional Signaling in a Complex World}},
volume = {25},
year = {2016}
}

@article{Williams1992,
author = {Williams, Ronald J},
journal = {Machine Learning},
number = {3-4},
pages = {229--256},
title = {{Simple statistical gradient-following algorithms for connectionist reinforcement learning}},
volume = {8},
year = {1992}
}

@book{Carroll1871,
author = {Carroll, Lewis},
publisher = {Macmillan},
title = {{Through the Looking-Glass, and What Alice Found There}},
year = {1871}
}

@inproceedings{Kingma2015,
archivePrefix = {arXiv},
arxivId = {1412.6980},
author = {Kingma, Diederik P. and Ba, Jimmy},
booktitle = {International {Conference} of {Learning} {Representations} ({ICLR})},
eprint = {1412.6980},
title = {{Adam: A Method for Stochastic Optimization}},
url = {https://arxiv.org/abs/1412.6980},
year = {2015}
}

@article{Barrett2009,
author = {Barrett, Jeffrey A},
doi = {10.1007/s11238-007-9064-0},
isbn = {1123800790},
journal = {Theory and Decision},
number = {2},
pages = {223--237},
title = {{The Evolution of Coding in Signaling Games}},
volume = {67},
year = {2009}
}

@article{Barrett2007a,
author = {Barrett, Jeffrey A},
journal = {Philosophy of Science},
pages = {527--546},
title = {{Dynamic Partitioning and the Conventionality of Kinds}},
volume = {74},
year = {2007}
}

@article{Kennedy2007,
author = {Kennedy, Christopher},
doi = {10.1007/s10988-006-9008-0},
journal = {Linguistics and Philosophy},
pages = {1--45},
title = {{Vagueness and grammar: the semantics of relative and absolute gradable adjectives}},
volume = {30},
year = {2007}
}

@inproceedings{Clevert2016,
author = {Clevert, Djork-Arn\'e and Unterthiner, Thomas and Hochreiter, Sepp},
booktitle = {International Conference of Learning Representations},
title = {{Fast and Accurate Deep Network Learning by Exponential Linear Units (ELUs)}},
url = {http://arxiv.org/abs/1511.07289},
year = {2016}
}

@article{Kennedy2005,
author = {Kennedy, Christopher and McNally, Louise},
journal = {Language},
number = {2},
pages = {345--381},
title = {{Scale Structure, Degree Modification, and the Semantics of Gradable Predicates}},
volume = {81},
year = {2005}
}

@inproceedings{Xu2015,
archivePrefix = {arXiv},
arxivId = {1502.03044},
author = {Xu, Kelvin and Ba, Jimmy and Kiros, Ryan and Cho, Kyunghyun and Courville, Aaron and Salakhutdinov, Ruslan and Zemel, Richard and Bengio, Yoshua},
booktitle = {International {Conference} on {Machine} {Learning} ({ICML} 32)},
editor = {Bach, Francis and Blei, David},
eprint = {1502.03044},
pages = {2048--2057},
title = {{Show, Attend and Tell: Neural Image Caption Generation with Visual Attention}},
url = {https://arxiv.org/abs/1502.03044},
year = {2015}
}

\appendix

\section{Full Experiment Details}

For each number of dimensions $n$, a context has $2n$ objects.  Each object is specified by $n$ real numbers, chosen uniformly at random from the interval $(0, 2)$ at steps of $0.1$.  The values are uniformly subtracted by $1$ to center them around $0$.  

The sender thus has $2n^2$ input nodes.  As a convention, the first object for the sender is always the target. It has two hidden layers of 64 nodes each, with exponential linear activation \cite{Clevert2016}.  The final hidden layer is then passed through two linear layers, with output sizes $| M_S |$ and $2$, respectively.  These are batch normalized \cite{Ioffe2015} and fed into a softmax, to generate distributions over $M_S$ and $M_P$.

The Basic Receiver receives the context, but with the objects in a random order compared to the sender, and two signals sampled from the sender's output distributions, encoded as one-hot vectors.  It then has three rectified linear hidden layers of 64, 64, and 32 units respectively.  Then a final linear layer with $2n$ output nodes (one for each target object) is passed through batch normalization and softmax to generate a distribution.

The Attentional Receiver passes the context and a messaged from $M_S$ sampled from the sender through one exponential linear layer of 64 units, before batch normalization and softmax of size $n$, one for each dimension.  A sample is taken from this distribution.  The corresponding scalar values for each object along the dimension, together with a message sampled from the sender's distribution over $M_P$ are passed through exponential linear layers of size 64 and 32, before batch normalization and softmax produce a distribution over target objects.

We trained using the REINFORCE algorithm, with mini-batches of size 64, and the Adam optimizer \cite{Kingma2015} with learning rate $5\cdot 10^{-4}$.  For $n = 1, 2, 3$ dimensions, and each type of receiver, we ran 10 trials of $5000$, $20000$, and $50000$ mini-batches of training.  After training, the trained networks then played $5000$ versions of the game; the signals chosen, the target chosen, whether it was correct, and what the `true' dimension and direction (min/max) for identifying the target in context were recorded.

Everything was implemented in PyTorch.  The code and data are available at \url{https://github.com/shanest/function-words-context}.

\end{document}